\documentclass[11pt,a4paper]{article}
\usepackage[margin=1in]{geometry}
\usepackage{url}
\usepackage[hidelinks]{hyperref}
\usepackage[utf8]{inputenc}
\usepackage[small]{caption}
\usepackage{graphicx}
\usepackage{amsmath}
\usepackage{amsthm}
\usepackage{booktabs}
\usepackage{algorithm}
\usepackage{algorithmic}
\usepackage{xcolor}
\usepackage{natbib}

\newcommand{\rd}[1]{\textcolor{red}{#1}}

\title{\textbf{Predicting missing values. A good idea?}}
\author{
    Stef van Buuren\\
    TNO Netherlands Organization for Applied Scientific Research, Leiden\\
    Dept.\ of Methodology \& Statistics, University of Utrecht\\
    \texttt{stef.vanbuuren@tno.nl}
}
\date{}

\begin{document}

\maketitle

\begin{abstract}

Minimizing the Mean Squared Error (MSE) is a key objective in machine learning and is commonly used for imputing missing values. While this approach provides accurate point estimates, it introduces systematic biases in downstream analyses. These biases affect key parameters such as variance, prevalence, correlation, slope, and explained variance. The root cause is that imputed values optimized for MSE are averages, which reduce the natural variability in the data.

This paper demonstrates that adding noise to imputed values can effectively eliminate these biases. The required noise level is proportional to the MSE. Using a toy example in a multivariate normal setting, we compare two methods: predictive imputation, which minimizes MSE, and stochastic imputation, which incorporates random noise. Simulation results show that predictive methods systematically introduce bias, while stochastic methods preserve the data’s natural variability and produce unbiased estimates.

We also evaluate three popular imputation tools — \texttt{missForest}, \texttt{softImpute}, and \texttt{mice} — and observe consistent biases in predictive methods. These findings highlight that MSE is an inadequate measure of imputation quality, as it prioritizes accuracy over variability. Incorporating noise into imputation methods is essential to prevent biases and ensure valid downstream analyses, underscoring the importance of stochastic approaches for handling incomplete data.

\end{abstract}

\section{Introduction}

Handling missing data is a fundamental challenge in data analysis and machine learning. Incomplete datasets are common in fields such as healthcare, finance, and social sciences, where the ability to accurately impute missing values is critical for downstream analyses. Among the various approaches, minimizing the Mean Squared Error (MSE) is widely regarded as a gold standard for imputation due to its focus on accuracy. This criterion has been extensively used in methods like regression-based imputation \citep{buck1960,gleason1975}, matrix factorization \citep{hastie2015}, and machine learning-based methods such as random forests \citep{stekhoven2011}. However, this focus on MSE comes with hidden costs that are often overlooked.

Minimizing MSE provides accurate point estimates but systematically reduces the natural variability in the data. As a result, biases are introduced in key downstream parameters, including variance, prevalence, correlation, slope, and explained variance. These biases can distort relationships in the data and lead to misleading conclusions in predictive modeling and decision-making. For instance, \citet{rubin1987} and \citet{schafer2000} demonstrated that deterministic imputation methods fail to account for the uncertainty inherent in missing data, resulting in overconfident estimates and invalid inferences.

Despite these limitations, MSE remains the dominant criterion for evaluating imputation methods in the machine learning community. Much of the existing literature focuses primarily on minimizing MSE \citep{troyanskaya2001,waljee2013,bertsimas2017,hegde2019,jarrett2022}, often neglecting its impact on the validity of downstream analyses. Recent work by \citet{ramosaj2022,shadbahr2023,sun2023} has highlighted these issues, but the broader implications for data variability and bias in downstream analyses remain underexplored.

This paper addresses these challenges by:
	1.	Demonstrating how minimizing MSE introduces systematic biases in downstream analyses.
	2.	Proposing the addition of noise to imputed values as a simple yet effective solution to preserve data variability.
	3.	Evaluating popular imputation tools, such as \texttt{missForest} \citep{stekhoven2011}, \texttt{softImpute} \citep{hastie2015}, and \texttt{mice} \citep{vanbuuren2011}, to identify their limitations and provide practical recommendations.

By shifting the focus from minimizing MSE to preserving the inherent variability of the data, we can ensure that imputation methods produce unbiased and reliable results for downstream analyses. This paper bridges the gap between statistical rigor and practical utility, offering a robust framework for handling incomplete data in machine learning.

\section{Model Definitions}

Handling missing data involves two key steps: (1) imputing the missing values using an
imputation model ($\mathcal{M}_\text{imp}$) and (2) analyzing the completed dataset
using a downstream model ($\mathcal{M}_\text{down}$). This section defines these
models and examines their roles in preserving the validity of downstream analyses.

\subsection{Imputation Model}

The imputation model generates replacement values for missing data ($y_\text{mis}$)
using fully observed predictors ($X$). The model can be expressed as:

\[
\mathcal{M}_\text{imp}: \quad y_{\text{mis}} = f(X; \beta) + \varepsilon,
\]

where $f(X; \beta)$ represents the predicted value based on the predictors and model
parameters, and $\varepsilon$ is the residual error term. For example, in a linear
regression imputation model, $f(X; \beta)$ could represent the predicted value of $y$
based on a set of regression coefficients ($\beta$).

The vector $y$ consists of $n_1$ observed values ($y_\text{obs}$) and $n_0$ missing
values ($y_\text{mis}$), where $n = n_1 + n_0$. The imputation model's goal is to
estimate the missing values in $y_\text{mis}$ based on $X$, which is fully observed.

\subsection{Downstream Model}

Once the missing values are imputed, the downstream model evaluates the completed
dataset ($\dot{y}$), which combines observed ($y_\text{obs}$) and imputed
($\dot{y}_\text{mis}$) values. The downstream model's goal is to estimate parameters
of interest (e.g., means, slopes, correlations) while accounting for variability
introduced during imputation.

For example, if $\dot{y}$ is treated as a response variable, the downstream model
can be expressed as:

\[
\mathcal{M}_\text{Y}: \quad \dot{y} = g(Z; \gamma) + \eta,
\]

where $g(Z; \gamma)$ is the predicted value based on downstream predictors $Z$, and
$\eta$ is the residual error term.

Alternatively, if $\dot{y}$ is treated as a predictor, the downstream model is given
by:

\[
\mathcal{M}_\text{X}: \quad x = g(\dot{y}, W; \delta) + \nu,
\]

where $x$ is the first column of $X$, $W$ is a matrix of additional predictors used
alongside $\dot{y}$, and $\nu$ captures the unexplained variability in $x$.

\subsection{Connecting the Models}

Once $y_\text{mis}$ is imputed using $\mathcal{M}_\text{imp}$, the completed dataset is passed to $\mathcal{M}_\text{down}$. This connection highlights the critical dependency of downstream analyses on the accuracy and variability of imputed values. To avoid issues with uncongeniality, we assume that $Z$ and $W$ are subsets of $X$.
\citep{meng1994}

For instance, consider a dataset where $y$ represents height, $X$ includes age and
gender, and some height values are missing. The imputation model could predict
missing heights using a regression equation based on $X$. The downstream model might
then analyze the completed dataset to study the relationship between height and
other variables, such as weight or socioeconomic status.

\begin{algorithm}[t]
    \caption{Imputation algorithm for a linear model with large sample size.}
    \label{alg:algorithm}
    \textbf{Input}: $y$, $X$ \\
    \textbf{Parameter}: method = \{\textit{predict}, \textit{draw}\} \\
    \textbf{Output}: $\dot{y}$
    \begin{algorithmic}[1]
        \STATE Split the data into observed and missing components:
        $y = \{y_\mathrm{obs}, y_\mathrm{mis}\}$ and
        $X = \{X_\mathrm{obs}, X_\mathrm{mis}\}$.
        \STATE Estimate the regression coefficients:
        \[
        \hat{\beta} = (X_\mathrm{obs}^\top X_\mathrm{obs})^{-1}
        X_\mathrm{obs}^\top y_\mathrm{obs}.
        \]
        \STATE Estimate the residual variance:
        \[
        \hat{\sigma}^2 =
        \frac{(y_\mathrm{obs} - X_\mathrm{obs} \hat{\beta})^\top
        (y_\mathrm{obs} - X_\mathrm{obs} \hat{\beta})}{n_1 - p - 1},
        \]
        where $n_1$ is the number of observed cases and $p$ is the number of predictors.
        \IF{method = \textit{draw}}
            \STATE Generate noise: sample $\dot{z} \sim \mathcal{N}(0, \hat{\sigma}^2)$,
            a vector of length $n_0$ (number of missing cases).
        \ELSE
            \STATE Set $\dot{z} = 0$ (no noise added).
        \ENDIF
        \STATE Impute missing values:
        \[
        \dot{y} = X_\mathrm{mis} \hat{\beta} + \dot{z}.
        \]
        \STATE \textbf{return} $\dot{y}$.
    \end{algorithmic}
\end{algorithm}

\subsection{Imputation Methods}

This section defines two methods to generate imputations: \textit{predict} and \textit{draw}.

The \textit{predict} method imputes the expected value, minimizing the mean squared error (MSE):
$$
\text{MSE} = \frac{1}{n_0} \sum_{i \in \text{mis}} \left( y_i - \dot{y}_i \right)^2,
$$
where $n_{\text{mis}}$ is the number of missing values, $y_i$ is the true value, and $\dot{y}_i$ is the imputed value, defined as
$$
\textit{predict:}\quad\dot{y}_i = \hat{y}_i,
$$
the predicted value $\hat{y}_i = f(X; \hat\beta)$ from the imputation model. By directly optimizing $\text{MSE}$, this method aims to provide the most accurate point estimate for the missing value based on the available data. The method is also known as \textit{regression imputation} or \textit{conditional mean imputation}. \citep{yates1933,little2020}

The second method, called \textit{draw}, builds on the first but adds noise to $\hat{y}_i$. The imputation is composed as
\[
\textit{draw:}\quad\dot{y}_i = \hat{y}_i + \varepsilon_i,
\]
where $\varepsilon_i$ is a random draw from the distribution of $\varepsilon$. This method acknowledges the inherent uncertainty about the hidden data and captures the distributional properties of $y$. \citep{rubin1978} However, note that \textit{draw} is suboptimal in the sense that
\[
\text{MSE}(\textit{draw}) \geq \text{MSE}(\textit{predict}).
\]

\subsection{Imputation Algorithm}

Algorithm~\ref{alg:algorithm} provides the basic algorithm for a simple linear model $f(X; \beta) = X\beta$.
This algorithm assumes that the sample size is sufficiently large to produce reliable
estimates of $\hat{\beta}$ and $\hat{\sigma}^2$. For small samples, additional steps
may be needed to account for the variability in these estimates. The algorithm can
also be adapted for other models, such as logistic regression or non-linear methods,
by modifying the regression function $f(X; \beta)$ and the noise distribution. \citep{vanbuuren2018}

\subsection{Example}

\begin{figure*}[tb]
    \centering
    \includegraphics[width=\textwidth]{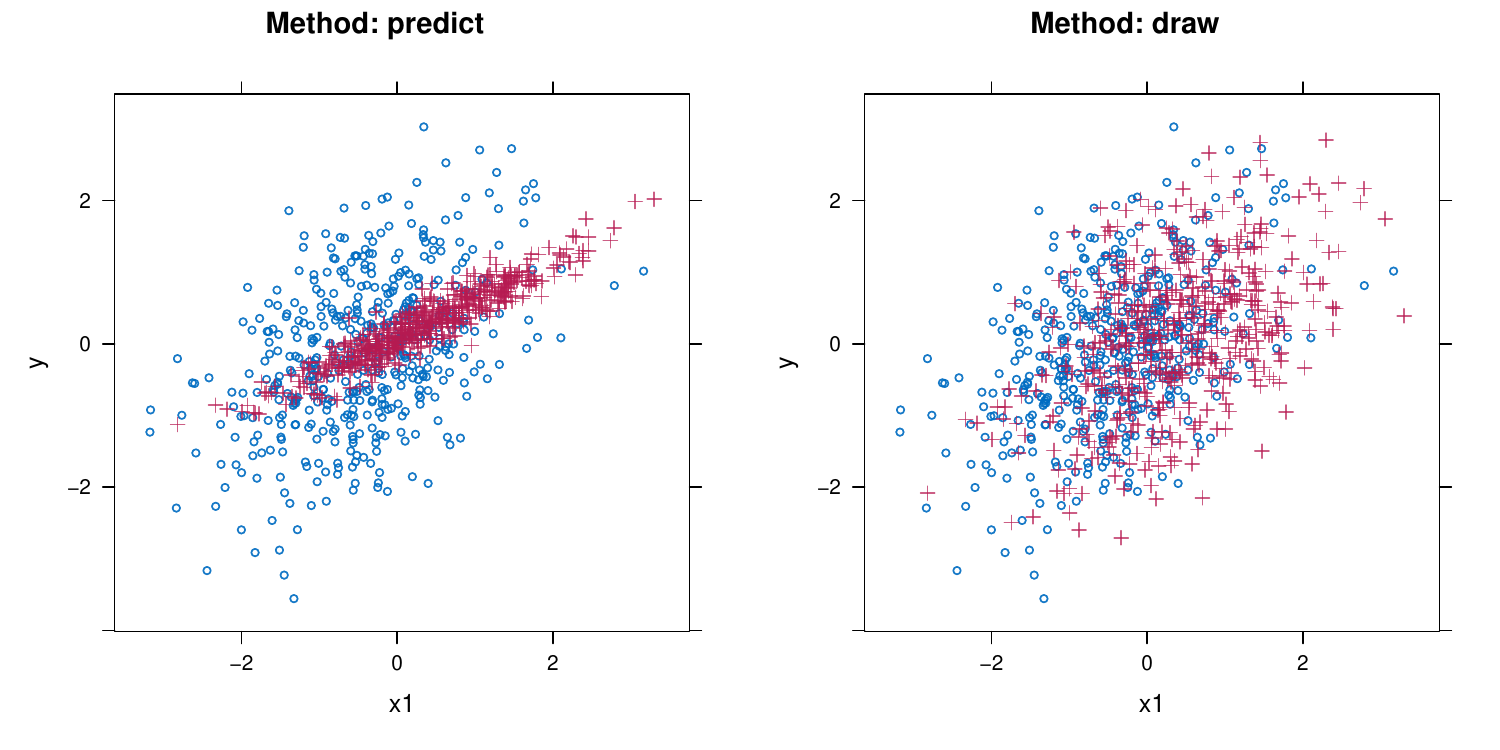}
    \caption{Visualization of two imputation methods applied to $y$ under a missing data mechanism with more missingness at higher $x_1$ values. Observed $y$-values are shown as blue circles, while red pluses indicate imputed values. The \textit{predict} method (left) produces deterministic imputations aligned with the conditional mean, while the \textit{draw} method (right) introduces variability by adding noise to better reflect the uncertainty of missing data.}
    \label{fig:example}
\end{figure*}

Figure~\ref{fig:example} illustrates the imputations generated by the \textit{predict} and \textit{draw} methods. The data were simulated from a bivariate normal distribution with $x_1$ and $x_2$ as predictors. A missing data mechanism was applied, where higher values of $x_1$ are associated with increased missingness in $y$. The \textit{predict} method imputes the expected value conditioned on $x_1$ and $x_2$, producing imputed values that lie on the regression plane defined by these predictors. In contrast, the \textit{draw} method adds random noise to the expected value, introducing variability that reflects the natural spread in the data.

\subsection{Research Questions}

This paper explores how the imputation model $\mathcal{M}_\text{imp}$ affects the validity of key estimates in the downstream model $\mathcal{M}_\text{down}$. Specifically, we study the impact of imputation techniques \textit{predict} and \textit{draw} on:
\begin{itemize}
    \item $\mu$: Mean of the imputed outcome $\dot{y}$,
    \item $\sigma$: Standard deviation of the imputed outcome $\dot{y}$,
    \item $P_{90}$: Percent cases exceeding the 90th percentile of $y$,
    \item $\rho$: Correlation between $\dot{y}$ and $x_1$,
    \item $\gamma$: Slope for $z_1$ in model $\mathcal{M}_\text{Y}$,
    \item $r_Y^2$: Proportion explained variance by model $\mathcal{M}_\text{Y}$,
    \item $\delta$: Slope for $\dot y$ in model $\mathcal{M}_\text{X}$,
    \item $r_X^2$: Proportion explained variance by model $\mathcal{M}_\text{X}$,
    \item $\text{MSE}$: Mean squared error of the imputed values.
\end{itemize}

If the data are Missing at Random (MAR), the downstream estimates of these parameters are expected to be similar to those in the original data.

\subsection{Expected effects}

For a normal linear model, the expected impact of the two imputation methods on the estimates of the parameters is as follows:

\begin{itemize}
    \item \textbf{$\mu$ (Mean):}
    \citet{schafer2000} demonstrated that both \textit{predict} and \textit{draw} are first-order approximations of the population mean. Consequently, we expect both methods to yield unbiased estimates of $\mu$.

    \item \textbf{$\sigma$ (Standard Deviation):}
    Imputed values from \textit{predict} are regressed towards the conditional mean, resulting in reduced variability compared to $y_\text{obs}$. Therefore, $\sigma$ is expected to be lower after \textit{predict}. Conversely, \textit{draw} adds noise to the imputed values, preserving the variability of the observed data.

    \item \textbf{$P_{90}$ (90th Percentile Proportion):}
    In the full dataset, the cutpoint is chosen such that 10\% of the cases lie above it. After \textit{predict}, this percentage will be lower due to the reduced spread of the conditional means. For \textit{draw}, the percentage is expected to closely match the population value.

    \item \textbf{$\rho$ (Correlation):}
    The \textit{predict} method tends to increase the correlation $r(\dot{y}, x_1)$ because imputations are drawn directly from the regression line. In contrast, \textit{draw} introduces appropriate variability, yielding a correlation close to the population value.

    \item \textbf{$\gamma$ (Slope for $x_1$):}
    Since the imputation and downstream models share the same outcome, the downstream model inherits the slope from the imputation model for both \textit{predict} and \textit{draw} imputation.

    \item \textbf{$r_Y^2$ (Proportion of Explained Variance for $\dot{y}$):}
    The \textit{predict} method increases $r_Y^2$ as it creates "perfect" predictions for the imputed values, leading to artificially high explained variance. By contrast, \textit{draw} is expected to produce values close to the population value.

    \item \textbf{$\delta$ (Slope for $\dot{y}$):}
    The slope $\delta$ is inflated under \textit{predict} due to the increased correlation ($\rho$). For \textit{draw}, the slope is expected to be unbiased.

    \item \textbf{$r_X^2$ (Proportion of Explained Variance for $x_1$):}
    Similarly, \textit{predict} inflates $r_X^2$ due to the higher correlation. The \textit{draw} method, however, is expected to return values that align with the population parameter.
\end{itemize}

\section{Simulation Study}

\begin{table}[tb]
\centering
\caption{Comparison of imputation methods across different signal levels and mechanisms. \textit{Color \rd{red} indicates problematic entries.}}
\resizebox{\textwidth}{!}{%
\begin{tabular}{lllrrrrrrrrr}
\hline
Signal & Method & Mechanism& $\mu$       & $\sigma$  & $P_{90}$ & $\rho$  &$\gamma$& $R_Y^2$  & $\delta$   & $R_X^2$  & MSE  \\
\hline
High   &        & Ground truth& -0.000   & 1.15      & 10.0     & 0.87    & 0.80   & 0.85     &     0.88   &     0.78 & 0.00 \\
       &\textit{predict}& MCAR&  0.000   & \rd{1.11} & \rd{9.1} &\rd{0.90}& 0.80   &\rd{0.92} & \rd{1.04}  & \rd{0.87}& 0.10 \\
       &            & MAR     & -0.004   & \rd{1.10} & \rd{8.9} &\rd{0.91}& 0.80   &\rd{0.92} & \rd{1.04}  & \rd{0.87}& 0.10 \\
       & \textit{draw} & MCAR & -0.000   & 1.15      & 10.0     & 0.87    & 0.80   & 0.85     &     0.88   &     0.78 & 0.20 \\
       &            & MAR     & -0.002   & 1.14      & 10.0     & 0.87    & 0.80   & 0.85     &     0.88   &     0.78 & 0.20 \\
\\
Low    &        & Ground truth& -0.001   & 1.04      & 10.0     & 0.48    & 0.40   & 0.26     &     0.33   &     0.35 & 0.00 \\
       &\textit{predict}& MCAR&  0.004   & \rd{0.82} & \rd{5.4} &\rd{0.60}& 0.40   &\rd{0.41} & \rd{0.57}  & \rd{0.42}& 0.40 \\
       &            & MAR     & -0.002   & \rd{0.83} & \rd{3.9} &\rd{0.60}& 0.40   &\rd{0.41} & \rd{0.57}  & \rd{0.42}& 0.40 \\
       & \textit{draw} & MCAR &  0.003   & 1.04      & 10.1     & 0.48    & 0.40   & 0.26     &     0.32   &     0.35 & 0.80 \\
       &            & MAR     & -0.003   & 1.04      & 10.1     & 0.48    & 0.40   & 0.26     &     0.32   &     0.35 & 0.80 \\
\hline
\end{tabular}}
\label{tab:imputation_comparison}
\end{table}

\subsection {Simulation setup}

We conducted a simulation study to test our expectations. We generated two synthetic populations, representing high-signal (\( R^2 = 0.8 \)) and low-signal (\( R^2 = 0.2 \)) data. Predictors \( x_1 \) and \( x_2 \) were drawn from a bivariate normal distribution with a correlation of 0.5. The outcome \( y \) was computed as a linear combination of \( x_1 \) and \( x_2 \) with variance contributions of 0.8 and 0.2, respectively, plus independent Gaussian noise to achieve the desired \( R^2 \). Each population consisted of \( N_\text{pop} = 10^6 \) observations.

Missing data were introduced only in the outcome variable \( y \) according to two mechanisms: MCAR (uniform missing data, independent of the predictors) and MAR (missingness is dependent on the predictor \( x_1 \), modeled using a right-censoring mechanism.). For both mechanisms, 50\% of the values in \( y \) are set to missing.

The sample size per replication was $N_\text{sample} = 1000$. For each sample, missing values in \( y \) were imputed using the two methods: \textit{predict} and \textit{draw} using the R methods \texttt{norm.predict} and \texttt{norm}, respectively.\citep{vanbuuren2011} The imputed datasets were then analyzed using the downstream models $\mathcal{M}_\text{Y}$ and $\mathcal{M}_\text{X}$, and the parameters of interest were estimated.

We set the number of replications to $T_\text{rep} = 200$, and averaged the estimates over the replications. For comparison, we also computed the population-level parameters for the complete data. The appendix contains the R script to generate the data and perform the analyses.

The simulation uses a simple setup that is unlikely to reflect real-world complexity. However, its simplicity allows precise control over the data-generating process and known true parameters. This provides a critical benchmark — if a method fails here, it is unlikely to succeed in more complex, realistic scenarios.

\subsection{Results}

Table~\ref{tab:imputation_comparison} summarizes the results for the high- and low-signal populations under MCAR and MAR mechanisms. The table shows that the \textit{predict} method systematically reduces the variance and the proportion of cases exceeding the 90th percentile. At the same time, it increases the correlation, the proportion of explained variance in both regressions, and the slope parameter. These biases are particularly pronounced in low-signal data. In contrast, the \textit{draw} method provides estimates that closely align with the true population values across all scenarios.

Notably, $\text{MSE}(\textit{draw})$ is twice as large as $\text{MSE}(\textit{predict})$. Based solely on the MSE criterion, one might incorrectly conclude that \textit{predict} is superior to \textit{draw}. However, the downstream estimates presented in Table~\ref{tab:imputation_comparison} clearly demonstrate that \textit{draw} should be the better method for downstream analyses. This highlights that relying exclusively on $\text{MSE}$ to evaluate and rank imputation methods can lead to misleading conclusions and poor method selection.

\section{Software comparison}

\subsection {Simulation setup}

We compared three widely used imputation methods: \textit{missForest} \citep{stekhoven2011}, \textit{softImpute} \citep{hastie2015}, and \textit{mice} \citep{vanbuuren2011}. \textit{missForest} and \textit{softImpute} are widely used imputation methods that represent predictive approaches focused on optimizing a loss function, such as minimizing reconstruction error or prediction error. These methods aim for accuracy by producing deterministic imputations aligned with their optimization criteria. In contrast, \textit{mice} is also widely used but adopts a fundamentally different approach, fully embracing randomness in its imputations through a Bayesian framework. Unlike predictive methods, \textit{mice} does not rely on a loss function, instead prioritizing the preservation of variability and uncertainty in the data. Comparing these paradigms provides valuable insights into their strengths and limitations across different scenarios.

Using the same synthetic data as in the simulation study, we applied these imputation methods to the missing values. The resulting imputed datasets were analyzed using the downstream models $\mathcal{M}_\text{Y}$ and $\mathcal{M}_\text{X}$ to estimate the parameters of interest. Apart from specifying a single imputation for \textit{mice}, all software packages were used with their default settings.

\begin{table}[tb]
\centering
\caption{Average parameter estimates from two downstream regression models applied to imputed datasets generated using three software packages (single imputation). \textit{Problematic entries are highlighted in \rd{red}.}}
\resizebox{\textwidth}{!}{%
\begin{tabular}{lllrrrrrrrrr}
\hline
Signal & Method         & Mechanism & $\mu$   & $\sigma$  & $P_{90}$ & $\rho$ & $\gamma$& $R_Y^2$ & $\delta$& $R_X^2$  & $\text{MSE}$ \\ \hline
High   &         &  Ground truth    & -0.002    &    1.15 &    10.0 &    0.87 &    0.80 &    0.85 &    0.88 &    0.78 & 0.00 \\
       & \textit{missForest} & MCAR &  0.000    &\rd{1.10}&\rd{ 9.1}&    0.89 &    0.78 &\rd{0.90}&\rd{1.01}&\rd{0.84}& 0.12 \\
       &                     & MAR  & -0.033    &\rd{1.07}&\rd{ 7.8}&    0.88 &\rd{0.75}&\rd{0.89}&\rd{1.03}&\rd{0.82}& 0.13 \\
       & \textit{softImpute} & MCAR &  0.001    &    1.15 &    10.0 &\rd{0.91}&\rd{0.84}&\rd{0.92}&\rd{0.99}&\rd{0.88}& 0.12 \\
       &                     & MAR  &  0.014    &    1.15 &    10.3 &\rd{0.91}&\rd{0.85}&\rd{0.92}&\rd{0.99}&\rd{0.88}& 0.10 \\
       & \textit{mice}       & MCAR &  0.001    &    1.15 &    10.0 &    0.87 &    0.79 &    0.85 &    0.88 &    0.78 & 0.20 \\
       &                     & MAR  & -0.015    &    1.13 &\rd{ 9.6}&    0.86 &    0.78 &    0.84 &    0.88 &    0.77 & 0.21 \\
\\
Low    &         &  Ground truth    & -0.000    &    1.04 &    10.0 &    0.48 &    0.40 &    0.26 &    0.33 &    0.35 & 0.00 \\
       & \textit{missForest} & MCAR & -0.004    &\rd{0.87}&\rd{ 5.7}&\rd{0.57}&    0.39 &\rd{0.37}&\rd{0.50}&\rd{0.40}& 0.48 \\
       &                     & MAR  & -0.009    &\rd{0.86}&\rd{ 4.7}&\rd{0.56}&    0.38 &\rd{0.36}&\rd{0.49}&\rd{0.39}& 0.48 \\
       & \textit{softImpute} & MCAR & -0.002    &\rd{1.73}&\rd{17.7}&\rd{0.43}&\rd{0.43}&\rd{0.52}&\rd{0.22}&\rd{0.57}& 2.79 \\
       &                     & MAR  &  0.039    &\rd{1.68}&\rd{18.5}&\rd{0.44}&\rd{0.45}&\rd{0.53}&\rd{0.24}&\rd{0.58}& 2.68 \\
       & \textit{mice}       & MCAR & -0.000    &    1.04 &    10.0 &    0.48 &    0.40 &    0.26 &    0.32 &    0.35 & 0.81 \\
       &                     & MAR  & -0.002    &    1.03 &    10.0 &    0.47 &    0.39 &    0.25 &    0.32 &    0.35 & 0.81 \\
\hline
\end{tabular}}
\label{tab:software-comparison}
\end{table}

\subsection{Results}

Table~\ref{tab:software-comparison} summarizes the results for the high- and low-signal populations under MCAR and MAR mechanisms. The table indicates that \textit{missForest} behaves similarly to the \textit{predict} method, showing consistent biases that become more pronounced as the relationships within the data weaken.

The behavior of \textit{softImpute} is less predictable. Under high-signal conditions, it performs well for $\sigma$ and $P_{90}$ but tends to overestimate the correlation and the proportion of explained variance. In the low-signal scenario, its imputed values exhibit excessive spread, leading to very high MSE. Additionally, \textit{softImpute} struggles to estimate $\gamma$, the slope of $x_1$ in model $\mathcal{M}_\text{Y}$, while consistently overestimating the proportion of explained variance. In contrast, \textit{mice} produces estimates that closely match the true population values across all cases. However, it exhibits slightly problematic behavior in the $P_{90}$ parameter. This issue can be resolved by using normal imputation instead of the default predictive mean matching.

As observed in previous analyses, the $\text{MSE}$ statistic misleadingly suggests that predictive methods outperform stochastic methods, highlighting the limitations of using $\text{MSE}$ as the sole criterion for evaluating imputation quality.

\section{Discussion}

\subsection{Key Findings}

This study shows key differences between predictive and stochastic imputation methods. Predictive methods, like \texttt{softImpute} and \texttt{missForest}, optimize accuracy, but reduce variability in imputed values, leading to biases in downstream analyses such as variance, correlations, and percentiles. Stochastic methods, like \texttt{mice}, add random noise to preserve variability, producing unbiased estimates. While predictive methods achieve lower MSE, this metric is misleading for evaluating imputation quality, as it overlooks the importance of variability. Biases are most pronounced in low-signal scenarios, where predictive methods overfit to the conditional mean, whereas stochastic methods better reflect the data’s underlying distribution.

\subsection{Previous work}

The idea of minimizing MSE for imputation is both intuitive and appealing. The approach is straightforward: generate data, introduce missingness, impute the missing values using a chosen method, and compare the imputed values to their true counterparts. The method with the lowest MSE wins. This simplicity makes it a tempting choice for researchers and aligns seamlessly with the machine learning paradigm, where optimizing loss functions is a core principle.

However, the limitations of MSE-based imputation methods have long been acknowledged in the statistical community. The third edition of \citet{little2020} outlines various formulas that quantify the biases introduced by predictive imputations. Similarly, \citet{schafer2000} proposed a method to correct variances for conditional mean imputation, but it remains largely unknown and rarely applied in practice.

This paper builds on the existing literature by systematically demonstrating the biases that arise in key downstream parameters, such as slopes, correlations, and explained variance, for models where $y$ serves as both an input and an output. The results are robust across different signal levels and missing data mechanisms, highlighting the generalizability of these findings. Additionally, the paper provides a description of the mechanisms driving these biases, emphasizing the critical role of preserving variability in imputed data to ensure valid downstream analyses, and suggests alternative criteria for evaluating imputation methods.

\subsection{Bias, variance and noise}

Missing values are, by definition, unknown, and it is generally impossible to recover them with complete accuracy. If precise recovery were possible, the values would not be missing in the first place. Predictive methods often assume that missing values can be accurately reconstructed, but this assumption is overly optimistic. A more pragmatic approach accepts the inherent uncertainty in missing data and explicitly accounts for it by marginalizing estimates over the variability inherent in the data.

In particular, the total MSE for the imputed data can be decomposed into three components \citep{prince2023}:
\[
\text{MSE}_\text{data} = \textit{Bias}^2 + \textit{Variance} + \textit{Noise}.
\]
Each component represents a distinct source of error. The \textit{Bias} is the squared difference between the average imputed value and the true value, typically arising from overly simplistic models that fail to capture the data's structure. The \textit{Variance} reflects the variability of imputed values across different realizations, often resulting from models that are too sensitive to noise in the observed data. The \textit{Noise} corresponds to the inherent randomness in the data that cannot be eliminated. For a missing value $y_i$, this irreducible error stems from the variability of $y_i$ given the predictors $X_i$. Mathematically, if $y_i$ is modeled as $y_i = f(X_i) + \varepsilon_i$, where $f(X_i)$ is the deterministic relationship between $X_i$ and $y_i$, and $\varepsilon_i$ is the random error term, then the irreducible error is $\sigma_\varepsilon^2 = \text{Var}(\varepsilon_i)$. This value is fixed for a given data-generating process and represents the variance of the random error term.

While bias and variance can be traded off, the noise is fixed. Predictive imputation methods like \textit{missForest}, which minimize $\text{MSE}$, reduce bias and variance but ignore the noise. These methods produce deterministic predictions, such as the conditional mean, lacking plausible variability. Attempts to compensate, such as the rescaling used in \textit{softImpute}, often fail to fully address this issue. In contrast, stochastic methods like \textit{mice} preserve the data’s natural variability, ensuring unbiased and valid downstream estimates.

\subsection{Effect of adding noise on the MSE}

How large should the noise be? The noise should be sufficient to capture the variability in the data but not so excessive that it overwhelms the imputed values. A common approach is to set the noise level equal to the residual variance of the imputation model, $\sigma_\varepsilon^2$. This ensures that the added noise accurately reflects the data’s inherent variability and remains consistent with the assumptions of the model.

What is the effect of adding noise to predictions on the magnitude of the MSE? Adding noise increases the MSE. The term $\text{MSE}(\textit{predict})$ reflects the residual error between the $n_0$ predictions and their hidden true values. The term $\sigma_\varepsilon^2$ is the residual variance of the imputation model with normally distributed errors, which is estimated using $n_1$ observed values in $y_\text{obs}$. Assuming that the residual variance estimated from the $n_1$ observed cases is used to impute the $n_0$ missing cases, we can sum the variances as $\text{MSE}(\textit{draw}) = \text{MSE}(\textit{predict}) + \sigma_\varepsilon^2$. Given that $\text{MSE}(\textit{predict}) \approx \sigma_\varepsilon^2$, it follows that $\text{MSE}(\textit{draw}) \approx 2 \cdot \text{MSE}(\textit{predict})$. This shows that adding noise approximately doubles the MSE.

\subsection{Alternative methods}

Several methodologies for evaluating imputation methods have been developed. \citet{rubin1987} proposed a set of frequentist criteria, bias, coverage, and confidence interval width, which should be evaluated in this order to ensure statistical validity. Diagnostics methods for comparing observed and imputed data \citep{abayomi2008,bondarenko2016,vanbuuren2018} are available. Frameworks developed by \citet{akande2017,morris2019,cai2023,shadbahr2023,oberman2024} assess the impact of imputation on downstream analyses in various ways. Among these approaches, posterior predictive checking \citep{gelman2004,cai2023} of the observed data offers a potential alternative to the MSE. It evaluates imputation quality in a manner that is agnostic to the downstream model, is versatile enough to work directly with incomplete data, and fits well in the machine learning paradigm.

Adding noise to the imputed values reduces their precision but is an important step forward to account for the inherent uncertainty of the missing data. For simplicity, this paper focused on single imputation. If the reduced precision is a concern, or if an estimate of the uncertainty in the downstream analysis is required, multiple imputation provides a solution to address both issues \citep{rubin1987,vanbuuren2018}.

\section{Conclusion}

Imputation methods should not be evaluated solely by their ability to recreate the true data. Instead, the goal of missing value imputation should be to facilitate informative and valid downstream analyses. Effective imputations must remain neutral, preserving the underlying relationships in the data without introducing distortions. It is crucial to recognize that imputation is not the same as prediction.

However, predicting missing values becomes a good idea when we move beyond simply optimizing the MSE and embrace the addition of noise to preserve data variability. This straightforward adjustment not only addresses the limitations of deterministic predictions but also ensures that downstream analyses remain valid and unbiased—a small yet powerful step toward more robust imputation methods.

\bibliographystyle{named}
\bibliography{references_arxiv}

\appendix

\section{R Code}

The following R scripts reproduce the tables and figure in this paper.
All scripts require R with the packages \texttt{MASS}, \texttt{mice}, \texttt{dplyr},
\texttt{missForest}, \texttt{softImpute}, \texttt{lattice}, and \texttt{gridExtra}.

\subsection{Simulation for Table~\ref{tab:imputation_comparison} (\texttt{simulate\_table1.R})}

{\small\begin{verbatim}
# Simulate properties of methods "predict" and "draw".
# Table 1 in the manuscript: Predicting missing values: A good idea?

library(MASS)   # 7.3-61
library(mice)   # 3.17.0
library(dplyr)  # 1.1.4

# Simulation parameters
set.seed(123)      # Ensure reproducibility
N_pop <- 1e6       # Population size
N_sample <- 1000   # Sample size per replication
T_rep <- 200       # Number of replications
missing_prop <- 0.5  # Proportion of missing values in y

generate_population <- function(r_squared, N, var_prop = c(0.8, 0.2)) {
  if (abs(sum(var_prop) - 1) > 1e-6) stop("var_prop must sum to 1.")
  Sigma <- matrix(c(1, 0.5, 0.5, 1), nrow = 2)
  predictors <- as.data.frame(mvrnorm(N, mu = c(0, 0), Sigma = Sigma))
  colnames(predictors) <- c("x1", "x2")
  beta1 <- sqrt(r_squared * var_prop[1])
  beta2 <- sqrt(r_squared * var_prop[2])
  predictors$y <- beta1 * predictors$x1 + beta2 * predictors$x2 +
    rnorm(N, mean = 0, sd = sqrt(1 - r_squared))
  return(predictors)
}

# Generate populations
high_signal_pop <- generate_population(r_squared = 0.8, N = N_pop)
low_signal_pop  <- generate_population(r_squared = 0.2, N = N_pop)

introduce_missingness <- function(data, mechanism = "MCAR",
                                  missing_prop = 0.5) {
  if (!"y" %in% colnames(data)) stop("Data must contain column 'y'.")
  patterns <- matrix(1, ncol = ncol(data), nrow = 1)
  colnames(patterns) <- colnames(data)
  patterns[1, "y"] <- 0
  if (mechanism == "MCAR") {
    weights <- rep(1, ncol(data))
  } else if (mechanism == "MAR") {
    weights <- c(1, 0, 0)
  } else {
    stop("Unsupported mechanism. Use 'MCAR' or 'MAR'.")
  }
  mice::ampute(data, patterns = patterns, mech = mechanism,
               type = "RIGHT", freq = c(1),
               prop = missing_prop, weights = weights)$amp
}

simulate <- function(pop_data, mechanism,
                     method_1 = "norm.predict", method_2 = "norm") {
  results <- data.frame()
  for (t in 1:T_rep) {
    sample_data    <- pop_data[sample(1:N_pop, N_sample), ]
    sample_missing <- introduce_missingness(sample_data,
                        mechanism = mechanism,
                        missing_prop = missing_prop)
    imputed_1 <- mice(sample_missing, method = method_1,
                      m = 1, maxit = 1, print = FALSE)
    imputed_2 <- mice(sample_missing, method = method_2,
                      m = 1, maxit = 1, print = FALSE)
    complete_1 <- complete(imputed_1, 1)
    complete_2 <- complete(imputed_2, 1)
    true_y     <- sample_data$y
    imputed_1_y <- complete_1$y
    imputed_2_y <- complete_2$y
    params <- data.frame(
      replication = t, mechanism = mechanism,
      method = c("predict", "norm"),
      mu    = c(mean(imputed_1_y), mean(imputed_2_y)),
      sigma = c(sd(imputed_1_y),   sd(imputed_2_y)),
      p90   = c(mean(imputed_1_y > quantile(true_y, 0.9)),
                mean(imputed_2_y > quantile(true_y, 0.9))),
      rho   = c(cor(complete_1$y, complete_1$x1),
                cor(complete_2$y, complete_2$x1)),
      gamma = c(lm(y ~ x1 + x2, data = complete_1)$coefficients[2],
                lm(y ~ x1 + x2, data = complete_2)$coefficients[2]),
      r2_y  = c(summary(lm(y ~ x1 + x2, data = complete_1))$r.squared,
                summary(lm(y ~ x1 + x2, data = complete_2))$r.squared),
      delta = c(lm(x1 ~ y + x2, data = complete_1)$coefficients[2],
                lm(x1 ~ y + x2, data = complete_2)$coefficients[2]),
      r2_x  = c(summary(lm(x1 ~ y + x2, data = complete_1))$r.squared,
                summary(lm(x1 ~ y + x2, data = complete_2))$r.squared),
      mse   = c(mean((true_y - imputed_1_y)^2),
                mean((true_y - imputed_2_y)^2))
    )
    results <- rbind(results, params)
  }
  return(results)
}

results_high_mcar     <- simulate(high_signal_pop, mechanism = "MCAR")
results_high_marright <- simulate(high_signal_pop, mechanism = "MAR")
results_low_mcar      <- simulate(low_signal_pop,  mechanism = "MCAR")
results_low_marright  <- simulate(low_signal_pop,  mechanism = "MAR")

final_results <- bind_rows(
  cbind(results_high_mcar,     signal = "high"),
  cbind(results_high_marright, signal = "high"),
  cbind(results_low_mcar,      signal = "low"),
  cbind(results_low_marright,  signal = "low")
)

summary_results <- final_results %>%
  group_by(signal, mechanism, method) %>%
  summarise(across(mu:mse, mean, na.rm = TRUE))

pop_high <- data.frame(signal = "high", mechanism = "", method = "",
  mu = mean(high_signal_pop$y), sigma = sd(high_signal_pop$y),
  p90 = mean(high_signal_pop$y > quantile(high_signal_pop$y, 0.9)),
  rho = cor(high_signal_pop$y, high_signal_pop$x1),
  gamma = lm(y ~ x1 + x2, data = high_signal_pop)$coefficients[2],
  r2_y  = summary(lm(y ~ x1 + x2, data = high_signal_pop))$r.squared,
  delta = lm(x1 ~ y + x2, data = high_signal_pop)$coefficients[2],
  r2_x  = summary(lm(x1 ~ y + x2, data = high_signal_pop))$r.squared,
  mse = 0)
pop_low <- data.frame(signal = "low", mechanism = "", method = "",
  mu = mean(low_signal_pop$y), sigma = sd(low_signal_pop$y),
  p90 = mean(low_signal_pop$y > quantile(low_signal_pop$y, 0.9)),
  rho = cor(low_signal_pop$y, low_signal_pop$x1),
  gamma = lm(y ~ x1 + x2, data = low_signal_pop)$coefficients[2],
  r2_y  = summary(lm(y ~ x1 + x2, data = low_signal_pop))$r.squared,
  delta = lm(x1 ~ y + x2, data = low_signal_pop)$coefficients[2],
  r2_x  = summary(lm(x1 ~ y + x2, data = low_signal_pop))$r.squared,
  mse = 0)

summary_results <- bind_rows(pop_high, summary_results[1:4, ],
                              pop_low,  summary_results[5:8, ])
ord <- c(1, 5, 3, 4, 2, 6, 10, 8, 9, 7)
print(summary_results[ord, ])
\end{verbatim}}

\subsection{Simulation for Table~\ref{tab:software-comparison} (\texttt{simulate\_table2.R})}

{\small\begin{verbatim}
# Simulate properties of methods "mice", "missForest", and "softImpute".
# Table 2 in the manuscript: Predicting missing values: A good idea?

library(MASS)       # 7.3-61
library(mice)       # 3.17.0
library(missForest) # 1.5
library(softImpute) # 1.4-1
library(dplyr)      # 1.1.4

set.seed(123)
N_pop <- 1e6;  N_sample <- 1000;  T_rep <- 200;  missing_prop <- 0.5

# (generate_population and introduce_missingness as in simulate_table1.R)

apply_imputation <- function(method, data) {
  if (method == "mice") {
    imputed <- mice(data, m = 1, print = FALSE)
    mice::complete(imputed, 1)
  } else if (method == "missforest") {
    missForest(as.data.frame(data), verbose = FALSE)$ximp
  } else if (method == "softimpute") {
    X <- as.matrix(data)
    fit <- softImpute(X)
    as.data.frame(softImpute::complete(X, fit))
  } else {
    stop("Unknown imputation method.")
  }
}

simulate <- function(pop_data, mechanism, methods) {
  results <- data.frame()
  for (t in 1:T_rep) {
    sample_data    <- pop_data[sample(1:N_pop, N_sample), ]
    sample_missing <- introduce_missingness(sample_data,
                        mechanism = mechanism,
                        missing_prop = missing_prop)
    for (method in methods) {
      imputed_data <- apply_imputation(method, sample_missing)
      true_y    <- sample_data$y
      imputed_y <- imputed_data$y
      params <- data.frame(
        replication = t, mechanism = mechanism, method = method,
        mu    = mean(imputed_y),
        sigma = sd(imputed_y),
        p90   = mean(imputed_y > quantile(true_y, 0.9)),
        rho   = cor(imputed_data$y, imputed_data$x1),
        gamma = lm(y ~ x1 + x2, data = imputed_data)$coefficients[2],
        r2_y  = summary(lm(y ~ x1 + x2, data = imputed_data))$r.squared,
        delta = lm(x1 ~ y + x2, data = imputed_data)$coefficients[2],
        r2_x  = summary(lm(x1 ~ y + x2, data = imputed_data))$r.squared,
        mse   = mean((true_y - imputed_y)^2)
      )
      results <- rbind(results, params)
    }
  }
  return(results)
}

high_signal_pop <- generate_population(r_squared = 0.8, N = N_pop)
low_signal_pop  <- generate_population(r_squared = 0.2, N = N_pop)
methods <- c("mice", "missforest", "softimpute")

results_high_mcar     <- simulate(high_signal_pop, "MCAR", methods)
results_high_marright <- simulate(high_signal_pop, "MAR",  methods)
results_low_mcar      <- simulate(low_signal_pop,  "MCAR", methods)
results_low_marright  <- simulate(low_signal_pop,  "MAR",  methods)

final_results <- bind_rows(
  cbind(results_high_mcar,     signal = "high"),
  cbind(results_high_marright, signal = "high"),
  cbind(results_low_mcar,      signal = "low"),
  cbind(results_low_marright,  signal = "low")
)

summary_results <- final_results %>%
  group_by(signal, mechanism, method) %>%
  summarise(across(mu:mse, mean, na.rm = TRUE))

ord <- c(5, 2, 6, 3, 4, 1, 11, 8, 12, 9, 10, 7)
print(summary_results[ord, ])
\end{verbatim}}

\subsection{Figure~\ref{fig:example} (\texttt{graph.R})}

{\small\begin{verbatim}
# Produce Figure 1: predict vs draw imputation on a single sample.

library(MASS);  library(mice);  library(dplyr)
library(lattice);  library(gridExtra)

set.seed(123)
N_pop <- 1e6;  N_sample <- 1000;  missing_prop <- 0.5

# (generate_population and introduce_missingness as in simulate_table1.R)

low_signal_pop <- generate_population(r_squared = 0.2, N = N_pop)

sample_data    <- low_signal_pop[sample(1:N_pop, N_sample), ]
sample_missing <- introduce_missingness(sample_data,
                    mechanism = "MAR", missing_prop = missing_prop)

predict <- mice(sample_missing, method = "norm.predict",
                m = 1, maxit = 1, print = FALSE)
draw    <- mice(sample_missing, method = "norm",
                m = 1, maxit = 1, print = FALSE)

p1 <- xyplot(predict, y ~ x1, main = "Method: predict",
             cex = c(0.6, 0.8), pch = c(1, 3))
p2 <- xyplot(draw,    y ~ x1, main = "Method: draw",
             cex = c(0.6, 0.8), pch = c(1, 3))

pdf(file = "graph.pdf", width = 10, height = 5, onefile = TRUE)
grid.arrange(p1, p2, nrow = 1)
dev.off()
\end{verbatim}}

\end{document}